\pdfoutput=1

\documentclass[11pt]{article}

\usepackage{acl}
\usepackage{times}
\usepackage{latexsym}

\usepackage[T1]{fontenc}

\usepackage[utf8]{inputenc}

\usepackage{microtype}

\usepackage{multirow}
\usepackage{multicol}
\usepackage{amsmath}
\usepackage{booktabs}
\usepackage{mathtools}
\usepackage{enumitem}
\usepackage{subcaption}
\usepackage{comment}
\usepackage{cleveref}

\usepackage{amsmath,amsfonts,bm}

\def\eqref#1{equation~\ref{#1}}

\def\1{\bm{1}}

\def\rvu{{\mathbf{i}}}

\def\rvu{{\mathbf{u}}}

\def\rvx{{\mathbf{x}}}
\def\rvy{{\mathbf{y}}}
\def\rvz{{\mathbf{z}}}

\DeclareMathAlphabet{\mathsfit}{\encodingdefault}{\sfdefault}{m}{sl}
\SetMathAlphabet{\mathsfit}{bold}{\encodingdefault}{\sfdefault}{bx}{n}

\definecolor{ourlightblue}{HTML}{C7EBFF}         
\definecolor{ourlightorange}{HTML}{FFDBC3}       
\definecolor{ourlightgray}{HTML}{EAEAF2}         
\definecolor{ourlightgreen}{HTML}{BDF0E2}        
\definecolor{ourdarkblue}{HTML}{0033CC}          

\title{Effective and Efficient Conversation Retrieval for \\ Dialogue State Tracking with Implicit Text Summaries}

\author{Seanie Lee$^\dagger$\thanks{* Work done during an internship at Apple.} \: Jianpeng Cheng$^\ddag$ \: Joris Driesen$^\ddag$ \: Alexandru Coca$^{\clubsuit *}$ \: Anders Johannsen$^\ddag$\\
$^\dagger$KAIST \: $^\ddag$Apple \: $^\clubsuit$University of Cambridge\\
  $^\dagger$\texttt{lsnfamily02@kaist.ac.kr} \\
  $^\ddag$\texttt{\{jianpeng.cheng, joris\_driesen, ajohannsen\}@apple.com} \\
  $^\clubsuit$\texttt{ac2123@cam.ac.uk}
  }

\begin{document}
\maketitle

\begin{abstract}

Few-shot dialogue state tracking (DST) with Large Language Models (LLM) relies on an effective and efficient conversation retriever to find similar in-context examples for prompt learning. Previous works use raw dialogue context as search keys and queries, and a retriever is fine-tuned with annotated dialogues to achieve superior performance. However, the approach is less suited for scaling to new domains or new annotation languages, where fine-tuning data is unavailable. To address this problem, we handle the task of conversation retrieval based on text summaries of the conversations.
A LLM-based conversation summarizer is adopted for query and key generation, which enables effective maximum inner product search. To avoid the extra inference cost brought by LLM-based conversation summarization, we further distill a light-weight conversation encoder which produces query embeddings without  decoding summaries for test conversations.  We validate our retrieval approach on MultiWOZ datasets with GPT-Neo-2.7B and LLaMA-7B/30B. The experimental results show a significant improvement over relevant baselines in few-shot DST settings.

\end{abstract}
\section{Introduction}\label{introduction}
Dialogue state tracking (DST) is one of the most crucial components in task-oriented dialogue systems. The goal of DST is to track users' intents, slots and values at every turn of a dialogue based on a predefined schema~\citep{multiwoz}. The challenge of training a supervised DST model lies in the cost of dialogue state annotations, which is not scalable to new schemas, domains or annotation languages. To address these challenges, recent works~\citep{ic-dst, smc-2} adopt in-context learning with pre-trained large language models (LLM) for few-shot DST. In the few-shot setting, similar dialogue exemplars are retrieved based on the test sample and then these exemplars are added to the LLM prompt for target generation. This approach is attractive since no domain-specific fine-tuning is required for the LLM but it  can still generalize to unseen domains.

One challenge in few-shot DST is how to retrieve salient conversation exemplars (e.g., in a set of 3 to 5) from the support set, which serves as demonstrations for the LLM. Ideally, a retrieved exemplar should carry both the same dialogue history and state change as the test sample. However, in a practical few-shot setting (e.g., with at most 100 annotated support examples), it is likely that no exemplar in the support set 
 satisfies the above requirement. Consider a test example with two user turns:
 
\begin{small}
\begin{verbatim}
user: book a flight to London Heathrow
system: where are you departing from
user: Amsterdam
\end{verbatim}
\end{small}

\noindent It is possible that the closest exemplar we can get from the support set is:

\begin{small}
\begin{verbatim}
user: I'm leaving Manchester by air
system: where are you flying to
user: To Paris
\end{verbatim}
\end{small}

\noindent  which  neither matches the test dialogue state nor the state change. Nevertheless,  we hope that LLM can generalize by learning from such exemplars with an identical user intent. The retrieval task gets harder when the conversation becomes lengthy with only partial history related to the current user input. For example, in another test dialog:

\begin{small}
\begin{verbatim}
user: what's the weather in London
system: sunny
user: book a flight to London Heathrow
system: where are you departuring from
user: Amsterdam
\end{verbatim}
\end{small}

\noindent the user's current intent is identical to the earlier test sample, but it involves unrelated history. Still we want to match the test sample to a similar exemplar, which reflects the user's intent up to the current point of conversation.
This retrieval cannot be easily accomplished with pre-trained dense retrieval models based on word or sentence similarity. To optimize retrieval performance, previous works~\citep{ic-dst, smc-2} fine-tune a dense retriever with ``structurally similar'' dialogue examples identified from dialogue state annotations with heuristics. \citet{ic-dst} additionally report that including dialogue state information in the retrieval key is helpful.  However, the approach is not  scalable to a practical few-shot setting (with fewer than 100 annotated support examples), as fine-tuning easily leads to overfitting and catastrophic forgetting~\citep{mccloskey1989catastrophic, seq-reptile}. It is also impractical to expect every domain owner to create their own fine-tuning data with well-engineered rules.

In this work, we propose a new solution for conversation retrieval starting with the introduction of a LLM-based conversation summarizer. For each exemplar to be indexed and also each test dialog, the summarizer produces a text summarizing \textit{what the user wants at this point of the conversation}. In Section \ref{overview}, we provide a discussion of this specific summarization choice and how it compares to dialogue state. The summaries are then used as condensed search keys and queries applicable to pre-trained dense retrievers with standard nearest neighbor search. We empirically show that in the few-shot setting, using summaries as retrieval keys and queries is more effective than using raw dialogues.

Notably the conversation summarization task described above can be easily handled by state-of-the-art LLMs via prompt learning, as we will show in an ablation study. However, the deployment of such a retrieval system also introduces extra model parameters and inference cost. Unlike search keys, which can be pre-built offline, a search query needs to be auto-regressively decoded for each test dialogue right during inference. To improve the efficiency of this conversation retriever, our second contribution in this work focuses on distilling a light-weight conversation encoder which embeds a raw dialogue directly into a vector space similar to the embedding of its summary. The light-weight conversation encoder enables efficient conversation search over a vector database without explicit query generation. When evaluated on the MultiWOZ dataset with GPT-Neo-2.7B~\citep{gpt-neo}, LLaMA-7B, and LLaMA-30B~\citep{llama} for few-shot DST, we find that the distilled conversation encoder is not only more efficient, but also more effective than a cascaded conversation retriever with explicit query generation. Our approach also significantly outperforms relevant baselines, which use annotated dialogues for retriever fine-tuning.

\section{Conversation Retrieval with Summaries for LLM-based DST \label{overview}}

In the context of  task-oriented dialogues with multiple turns of interactions between a user and a system, the objective of DST is to predict the accumulated intents, slots and values at each user turn. In a LLM-based approach, the generation of a dialogue state is conditioned on a task-specific prompt. The prompt includes at least the test conversation and a set of $k$ demonstration examples, from which we expect the LLM to learn to generalize. Considering the size limit of the prompt, $k$ is expected to be small (3-5 examples). Each of the retrieved examples is an annotated conversation sharing similar features as the test conversation. We expect to retrieve these exemplars with a dense retriever from a ``support set'' (e.g., 100 annotations) that can be constructed with minimum effort for domain scaling. 

There are two major challenges of conversation retrieval for LLM-based DST described above. First, a good representation of search keys and queries need to be found. As we analyze in Section \ref{introduction}, the similarity of two dialogues is not directly quantifiable by semantic distance, but rather requires more sophisticated structural matching mechanism or a higher-order similarity function. This requirement leads to the second challenge as to how to train an effective conversation retriever that can scale across domains. Previous works~\citep{ic-dst, smc-2} mainly fine-tune pre-trained dense retrievers with annotated dialogues obtained from the support set. However, fine-tuning is not realistic in a few-shot setting and for every domain.

\begin{figure}
    \centering
    \includegraphics[width=1.0\linewidth]{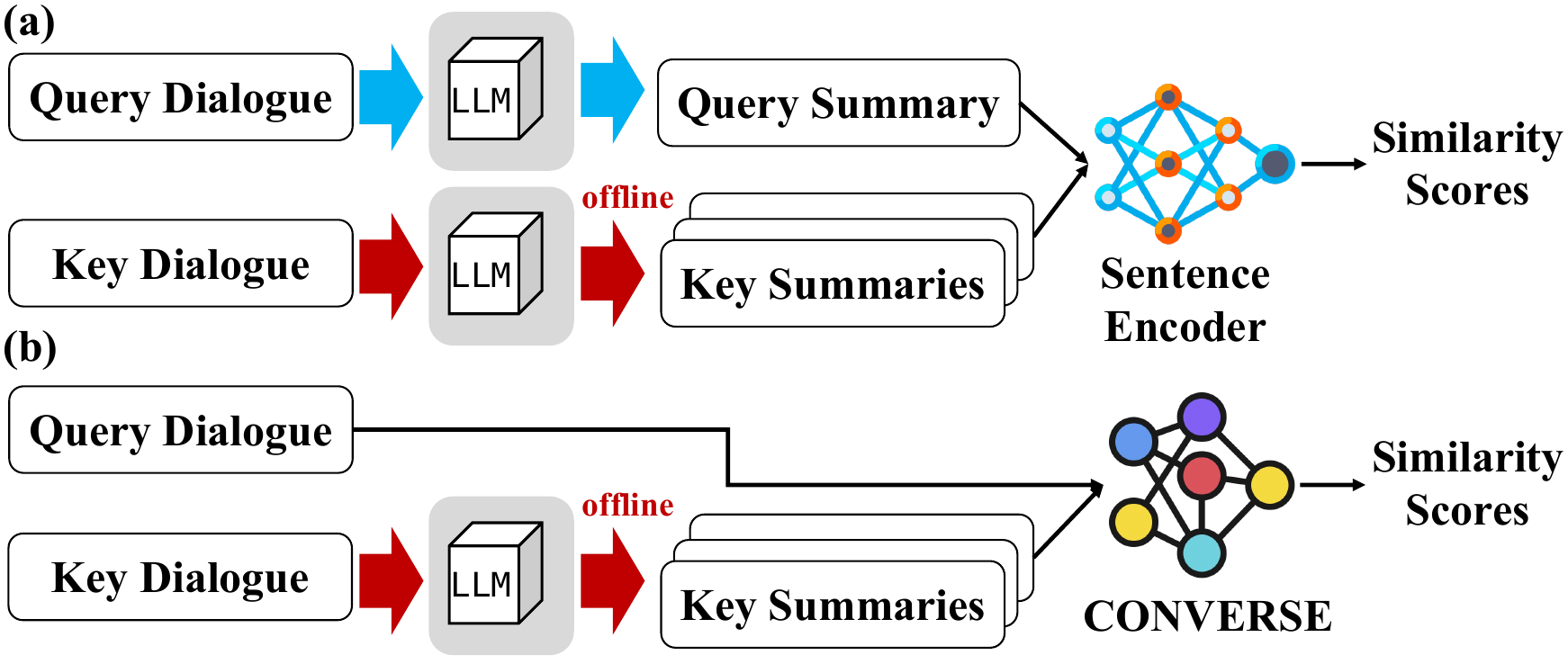}
    \vspace{-0.22in}
    \caption{\small Comparison between \textbf{(a)} off-the-shelf retriever with query generation and \textbf{(b)} CONVERSE w/o query generation.}
    \vspace{-0.2in}
    \label{fig:comp}
\end{figure}

As shown in Figure~\ref{fig:comp}-(a), our work introduces a query/key generation step in the LLM-based DST. The generation is performed with another LLM which transforms the raw dialogue context into a text summary whose similarity can be evaluated more easily with pretrained retrievers. Specifically, the text summary represents the user's intent up to the current point of the conversation. It grounds the latest user input onto the dialogue history, keeping only information related to the current user intent. Note that the summary is a contextual rewriting of the current user intent that is possibly expressed in multiple turns, with applied ellipsis recovery~\citep{ellipsis} and co-reference resolution~\citep{coref}. Examples of the conversation summary are in Table~\ref{tab:summary-1}. The summary can also be viewed as a text description of an updated dialogue state which is to be predicted by the LLM. Unlike the dialogue state, the summary does not maintain all conversation history but only includes information relevant to the current user input.

With the introduction of an explicit query/key generation step, we expect that the conversation retrieval becomes easier and the search index can be built more efficiently. To construct the search index, an offline process can be triggered to generate text summaries for every example in the support set. Note that search key generation does not add any inference cost. However, the query generation step comes at an extra cost since the generation needs to happen in an online process. In the next section, we describe how to make the conversation retrieval more efficient by stepping away from explicit query generation. 
\begin{figure*}
    \centering
    \includegraphics[width=1.0\linewidth]{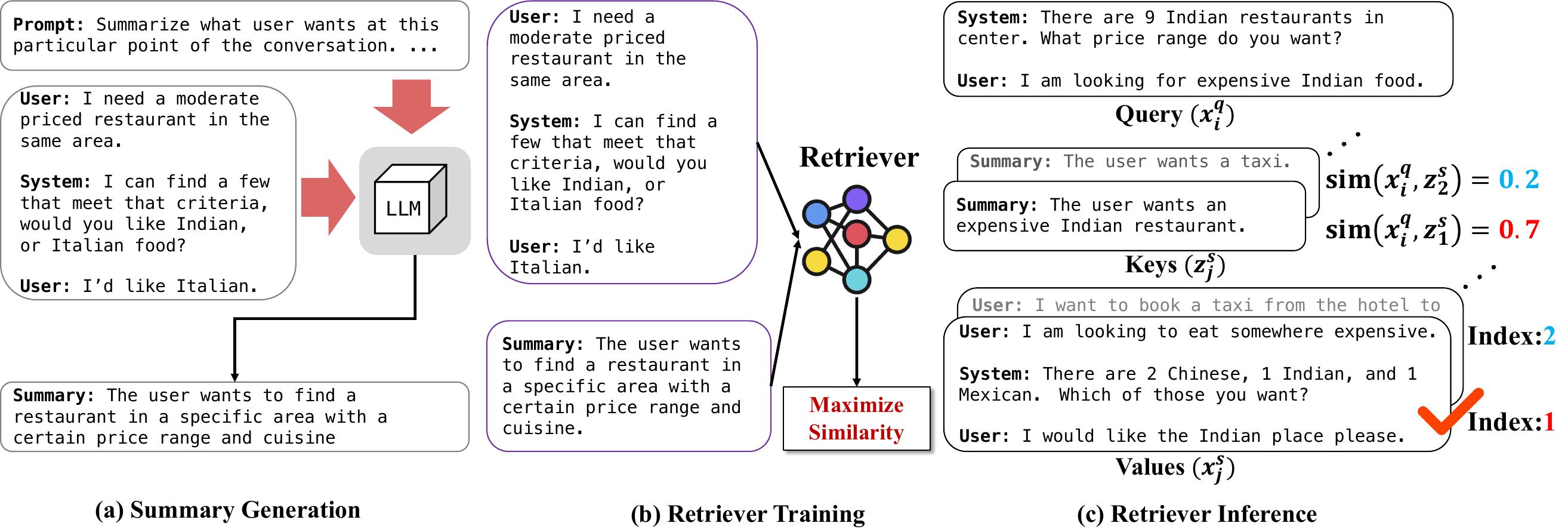}
    \vspace{-0.25in}
    \caption{\small\textbf{Concept.} \textbf{(a)} Generating a summary of a dialogue with language model (LM). \textbf{(b)} Training the retriever to maximize a similarity between the dialogue and generated summary. \textbf{(c)} Given a test dialogue as a query, we retrieve the dialogue (value) of which summary (key) obtains the best similarity score with the query.}
    \label{fig:concept}
    \vspace{-0.15in}
\end{figure*}
\section{Conversation Encoder Distillation}\label{sec:method}
Note that in the proposed conversation retriever, the LLM-based conversation summarizer needs to be invoked for every test sample to generate the search query as shown in Figure~\ref{fig:comp}-(a). To eliminate the extra inference cost, we propose to distill a light-weight conversation encoder which directly embeds a dialogue into a vector space similar to its summary, by maximizing their embedding similarity. The encoder is trained with large-scale dialogue-summary pairs generated by the conversation summarizer in an offline process. After training the model, as shown in Figure~\ref{fig:comp}-(b), we can directly encode each dialogue into a query embedding for maximum inner product search.  We call our conversation encoder CONVERSE, standing for  \textbf{CON}versation embeddings for \textbf{VE}rsatile \textbf{R}etrieval with implicit \textbf{S}ummari\textbf{E}s. Next we explain the structure and training objective of CONVERSE.

\subsection{Model}
\vspace{-0.05in}
\paragraph{Preliminaries} In our problem setup, we are given a set of unlabeled conversations between a user and system, denoted as $\mathcal{D}^u=
\{\rvx_i \}_{i=1}^n$ where each conversation $\rvx_i$ consists of $l_i$ utterances $(\rvu_{i, j}, \ldots, \rvu_{i, l_i})$ and each utterance $\rvu_{i,j}$ is a sequence of  $T_{i,j}$ tokens $(x_{i,j,1}, \ldots, x_{i,j, T_{i,j}})$. As shown in Figure~\ref{fig:concept}-(b), the training data of CONVERSE is prepared by invoking the conversation summarizer to generate a summary for each conversation $\rvx_i$, denoted as $\rvz_i$, which consists of $T^\prime_i$ tokens with $T^\prime_i \ll \sum_{j=1}^{l_i} T_{i,j}$. We denote the dataset augmented with summaries as $\mathcal{D}^a=\{(\rvx_i,\rvz_i) \}_{i=1}^n$. For brevity, we omit the first subscript $i$ if there is no ambiguity. Given the set of conversation-summary pairs, the goal is to train an encoder $f_\theta: \mathcal{V}^T \to \mathbb{R}^{T\times d}$ such that the similarity between a conversation and its summary is maximized, where $\mathcal{V}$ denotes a set of predefined tokens.

\vspace{-0.05in}
\paragraph{Conversation and Summary Embedding}
To match a conversation against a summary, we leverage the commonly used architecture in dense retrieval known as the dual encoder~\citep{dual1,dual2,dual3}, where a conversation and a summary are encoded jointly for similarity comparison. State-of-the-art dual encoders ~\citep{colbert} represent each encoding as multiple vectors, typically the contextualized token vectors, to represent the text. These models largely improve the model expressiveness, and exhibit much stronger performance and robustness compared to their single-vector counterparts~\citep{beir}. Based on it, we represent both the conversation embedding $f_\theta(\rvx)$ and the summary embedding $f_\theta(\rvz)$ as a matrix. While the summary encoder in the dual architecture can be directly integrated into off-the-shelf sentence encoders, our conversation encoder (CONVERSE) is designed to reflect the inductive bias of the summarization task.

\vspace{-0.05in}
\paragraph{CONVERSE}
Remember that the task of the conversation summarizer is to summarize the current user intent by grounding it to the conversation history. Hence the latest user input (the state delta) is most important and any past utterances irrelevant to the latest input should be dropped out.

To reflect the nature of the summarization task, we explicitly model the grounding step between the latest user input $\rvu_l$ and past utterances $\rvu_1, \ldots, \rvu_{l-1}$ as a structural bias in CONVERSE. This is achieved with the introduction of a soft retrieval structure that softly retrieves past utterances or tokens which are relevant to the latest user input. Specifically, the soft retrieval is simulated with another neural network $g_\phi: \mathbb{R}^d\times \mathbb{R}^d \to [0,1]$, which outputs the relevance score of each token in the utterances $\rvu_1, \ldots, \rvu_{l-1}$ conditioned on the latest user utterance $\rvu_l$. Then, the relevance scores are used to downweight irrelevant token representations of the conversation $\rvx$: 
\begin{gather}
    \hat{f}_{\theta, \phi}(\rvx) = \begin{bmatrix}
    w_{1,1} f_\theta(\rvx)^\top_{1,1}  \\
    \vdots \\
    w_{l-1,T_{l-1}} f_\theta(\rvx)^\top_{l-1,T_{l-1}} \\
    f_\theta(\rvx)^\top_{l, 1}  \\
    \vdots \\
    f_\theta(\rvx)^\top_{l, T_l}
    \end{bmatrix} \in \mathbb{R}^{T \times d} \nonumber \\
    w_{j,t} = g_\phi (f_\theta(\rvx)_{j,t}, s_l(\rvx)) \in [0,1] \label{eq:weight}\\
        s_l(\rvx) = \frac{1}{T_l} \sum_{t=1}^{T_l} f_\theta(\rvx)_{l, t} \in \mathbb{R}^d, \nonumber
\end{gather}
where $t\in \{1,\ldots, T_j\}$ for each $j\in \{1,\ldots, l-1\}$ and $f_\theta(\rvx)_{j,t}$ is a contextual representation of $t$-th token in $\rvu_j$. Intuitively, an irrelevant token in the conversation history receives a small weight, reducing its contribution to the final similarity scoring against the summary. Conversely, a token in the latest user input always carries the highest weight 1 and contributes more to the similarity computation.

\subsection{Training Objective}
\label{sec:3.2}

Given the conversation encoder $\hat{f}_{\theta, \phi}$ and summary encoder $f_\theta$ with a set of conversation and summary pairs $\mathcal{D}^a=\{(\rvx_i, \rvz_i) \}_{i=1}^n$, as illustrated in Figure~\ref{fig:concept}-(b), we train the dual encoder to maximize the similarity between a dialogue and its summary with the contrastive loss~\citep{henderson-cont}:
\begin{gather}
     L(\theta, \phi; \mathcal{D}^a)=-\frac{1}{n}\sum_{(\rvx,\rvz)\in\mathcal{D}^a} \log p_{\theta,\phi}(\rvz|\rvx) \label{eq:obj}\\
     p_{\theta,\phi}(\rvz |\rvx) = \frac{\exp(\texttt{sim}(\hat{f}_{\theta, \phi}(\rvx), {f}_\theta(\rvz)))}{\exp(\sum\limits_{\mathclap{(\rvx^\prime, \rvz^\prime)\in \mathcal{D}^a}}\texttt{sim}(\hat{f}_{\theta,\phi}(\rvx^\prime), f_\theta(\rvz^\prime)))}, \nonumber
\end{gather}
where $\texttt{sim}$ is the multi-vector similarity function~\citep{colbert},
which computes the similarity between the conversation and its summary, denoted as $\texttt{sim}(\hat{f}_{\theta,\phi}(\rvx), f_\theta(\rvz))$, by averaging maximum dot product between summary tokens and each conversation token as:
\begin{align}
         \frac{1}{T} \sum_{j=1}^l\sum_{t=1}^{T_j} \max_{t^\prime \in \{1,\ldots, T^\prime\}}\hat{f}_{\theta,\phi}(\rvx)_{j,t}^\top f_\theta(\rvz)_{t^\prime}.
\label{eq:colbert}
\end{align}
In practice, due to  computational costs, we sample a mini-batch $\mathcal{B}\subset\mathcal{D}^a$ for computing the denominator of  the contrastive loss in~\eqref{eq:obj}.

\subsection{Inference}
In LLM-based DST, we are given a small support set of labeled dialogues $\mathcal{D}^s_1=\{(\rvx^s_{i}, \rvy^s_i) \}_{i=1}^m$.
The search keys can be pre-built offline by calling the conversation summarizer to generate a summary for each dialogue $\rvx^s_i$ from the support set $\mathcal{D}^s_1$, resulting in a set of (conversation, label, and summary) triplets denoted as $\mathcal{D}^s_2 = \{(\rvx^s_i, \rvy^s_i, \rvz^s_i)\}_{i=1}^m$. The search index is then built with the summary as the key, and  a labeled conversation as the value. The summaries are encoded with the fine-tuned summary encoder described in Section~\ref{sec:3.2}.

During inference, for each conversation $\rvx^q_i$ from the test set $\mathcal{D}^q = \{\rvx^q_i\}_{i=1}^Q$,  we embed the conversation with the CONVERSE encoder and compute its similarity with every search key using the similarity function in~\eqref{eq:colbert},  i.e.,  $\texttt{sim}(\hat{f}_{\theta, \phi}(\rvx^q_i), f_\theta(\rvz^s_j))$ for $j=1,\ldots, m$. As shown in Figure~\ref{fig:concept}-(c), the retriever ranks examples $(\rvx^s_j, \rvy^s_j)$ based on the similarity score  and chooses the top-$k$ exemplars. Finally, the retrieved exemplars are added to the prompt of the downstream LLM for dialog state generation.

\section{Experiments}
\subsection{Experimental Setup}
\paragraph{Common}
We evaluate LLM-based DST with the proposed conversation retriever on MultiWOZ 2.1~\citep{multiwoz-2.1} and 2.4~\citep{ye2022multiwoz-2.4}. 

To simulate few-shot scenario, we consider a support set of 100 labeled conversations as the default setting in our comparison. For each experimental run, we randomly sample 100 labeled conversations from the training data of MultiWOZ 2.1/2.4. The analysis of other support set sizes is deferred to an ablation study. During inference, we retrieve the top 5 examples from the support set. The examples along with a test conversation are inserted into the prompt, following~\citet{ic-dst}. This setting is applied to all comparisons. We use both GPT-Neo~\citep{gpt-neo} and LLaMA-7B/30B~\citep{llama} as the LLM for DST generation.  For evaluation, we report average and standard deviation of Joint Goal Accuracy (JGA) and F1 score~\citep{henderson2014second} on all 7,368 test dialogues from MultiWOZ with three runs.

\paragraph{Baselines}
We compare the proposed conversation retriever with the following baselines.

\begin{enumerate}[itemsep=0mm, parsep=0pt, leftmargin=*]
\item \textsc{IC-DST}~\citep{ic-dst}: It utilizes dialogue labels to construct positive and negative pairs for fine-tuning a pretrained SBERT~\citep{sbert} or LinkBERT~\citep{linkbert} as a retriever. The retrieval key is a dialogue context, and the best dialogue context is reported to be previous dialogue state + current user input (which is better than a full dialogue).

\item \textsc{SM2}~\citep{smc-2}: Similar to IC-DST, it fine-tunes SBERT on labeled dialogue data with contrastive loss, where conversations with partial matching slots or values are considered as positive samples. The retrieval key is a dialogue context similar to IC-DST.

\item \textsc{GTR-T5-Large}~\citep{t5-gtr}: It uses a  T5 encoder, which is pretrained on large scale  corpora for sentence representation, to compute the similarity between conversations for retrieving examples. The retrieval key is the full dialogue.

\item \textsc{Jina-Large}~\citep{jina}: Similar to GTR-T5, the pretrained sentence encoder Jina is used to compute similarity between conversations. The retrieval key is also the full dialogue.

\end{enumerate}

\paragraph{Ours}
We use \texttt{gpt-3.5-turbo}~\citep{ChatGPT} as the conversation summarizer, since it provides reliable summaries that satisfy the task requirement in the prompt (see human evaluation in \ref{Qualitative} and the prompt specified in Appendix~\ref{app:prompt}). First, we evaluate the effectiveness of summary-based search key and query generation, using off-the-shelf retrievers GTR-T5-Large and Jina-Large, which are directly comparable with the baseline.

Second, we evaluate the distilled conversation encoder (CONVERSE). 
To train CONVERSE, we use the same conversation summarizer to generate a summary for every turn of every conversation from the full MultiWOZ training set, resulting in a total of 56,776 conversation-summary pairs. The parameters $\theta$ of the dual encoder $f_\theta$ and $\hat{f}_{\theta, \phi}$ are shared and initialized with LinkBERT~\citep{linkbert}, and trained on the conversation-summary pairs for 20 epochs with the objective in~\eqref{eq:obj}. 
LinkBERT~\citep{linkbert} is chosen since we empirically find that it offers the best general-purpose weight of initialization. We use the AdamW optimizer~\citep{adamw} with learning rate $5\cdot 10^{-5}$ and batch size 200. We use eight A100 GPUs for training the model.

\subsection{Quantitative Results}
\begin{table}[t]
\centering
\resizebox{0.48\textwidth}{!}{\begin{tabular}{lllll}
\toprule
                       & \multicolumn{2}{c}{\textbf{MultiWOZ 2.1}}                 & \multicolumn{2}{c}{\textbf{MultiWOZ 2.4}}                 \\
\cmidrule(lr){2-3} \cmidrule(lr){4-5}
Model                  & \multicolumn{1}{c}{JGA} & \multicolumn{1}{c}{F1} & \multicolumn{1}{c}{JGA} & \multicolumn{1}{c}{F1} \\
\midrule[0.8pt]
\multicolumn{5}{c}{GPT-Neo 2.7B~\citep{gpt-neo}}                                                                                             \\
\midrule[0.8pt]
IC-DST (SBERT)                          & $6.76$\tiny$\pm0.87$           & $42.91$\tiny$\pm2.87$           & $6.81$\tiny$\pm1.05$          & $43.42$\tiny$\pm3.18$         \\
IC-DST (LinkBERT)                          & $6.39$\tiny$\pm1.72$           & $40.11$\tiny$\pm3.30$           & $6.35$\tiny$\pm1.14$          & $40.78$\tiny$\pm3.10$         \\

SM2 & $5.44$\tiny$\pm0.27$ & $35.15$\tiny$\pm1.80$ & $5.33$\tiny$\pm0.76$  & $35.03$\tiny$\pm1.42$\\
GTR-T5                    & $4.77$\tiny$\pm0.66$           & $28.58$\tiny$\pm0.79$           & $4.66$\tiny$\pm0.57$          & $28.50$\tiny$\pm0.84$          \\
Jina                      & $5.11$\tiny$\pm0.18$           & $30.93$\tiny$\pm1.29$           & $5.16$\tiny$\pm0.40$          & $30.84$\tiny$\pm1.33$          \\
\midrule[0.5pt]
Sum. + GTR-T5          & $6.16$\tiny$\pm0.54$           & $40.60$\tiny$\pm2.51$           & $6.01$\tiny$\pm0.60$          & $40.40$\tiny$\pm2.34$         \\

Sum. + Jina            & $6.09$\tiny$\pm0.71$           & $40.48$\tiny$\pm2.62$           & $6.13$\tiny$\pm0.77$          & $40.84$\tiny$\pm2.95$          \\
\midrule[0.5pt]
\textbf{CONVERSE}           & $\textbf{8.07}$\tiny$\pm0.62$  & $\textbf{44.11}$\tiny$\pm2.45$  & $\textbf{7.85}$\tiny$\pm0.65$ & $\textbf{44.92}$\tiny$\pm2.16$                       \\
\midrule[0.8pt]
\multicolumn{5}{c}{LLaMA-7B~\citep{llama}}                                                                                            \\
\midrule[0.8pt]
IC-DST (SBERT)                          &   $18.30$\tiny$\pm2.81$            &   $69.51$\tiny$\pm3.36$       & $18.57$\tiny$\pm3.17$                 &   $70.37$\tiny$\pm3.54$                     \\
IC-DST (LinkBERT)                          &   $18.09$\tiny$\pm0.08$            &   $69.41$\tiny$\pm0.65$       & $18.97$\tiny$\pm0.53$                 &   $70.29$\tiny$\pm0.59$                     \\
SM2 &$15.23$\tiny$\pm1.56$ & $64.36$\tiny$\pm2.36$ & $15.01$\tiny$\pm1.72$ & $65.12$\tiny$\pm2.36$\\
GTR-T5                    &   $13.64$\tiny$\pm0.16$            &   $57.95$\tiny$\pm0.46$       & $13.61$\tiny$\pm0.43$                 &   $58.26$\tiny$\pm0.44$                     \\
Jina                      &    $15.58$\tiny$\pm0.58$           &   $60.89$\tiny$\pm0.41$       & $15.50$\tiny$\pm1.02$                  &   $61.48$\tiny$\pm0.34$                      \\
\midrule[0.5pt]
Sum. + GTR-T5          &    $17.54$\tiny$\pm0.34$           &   $68.36$\tiny$\pm0.48$       & $17.74$\tiny$\pm0.68$                 &   $69.14$\tiny$\pm0.77$                     \\
Sum. + Jina            &    $17.85$\tiny$\pm0.41$           &   $68.70$\tiny$\pm0.46$           & $18.37$\tiny$\pm0.61$             &   $69.65$\tiny$\pm0.87$                     \\
\midrule[0.5pt]
\textbf{CONVERSE}           &    $\textbf{19.33}$\tiny$\pm0.91$  &   $\textbf{71.48}$\tiny$\pm1.50$  & $\textbf{20.35}$\tiny$\pm1.03$    &   $\textbf{72.45}$\tiny$\pm1.52$                     \\
\bottomrule
\end{tabular}
}
\vspace{-0.1in}
\caption{\small JGA and F1 using labeled 100 conversations with GPT-Neo-2.7B and LLaMA-7B.}
\label{tab:main-exp}
\vspace{-0.1in}
\end{table}
\begin{table}[t]
\centering
\resizebox{0.48\textwidth}{!}{\begin{tabular}{lllll}
\toprule
                       & \multicolumn{2}{c}{\textbf{MultiWOZ 2.1}}                 & \multicolumn{2}{c}{\textbf{MultiWOZ 2.4}}                 \\
\cmidrule(lr){2-3} \cmidrule(lr){4-5}
Model                  & \multicolumn{1}{c}{JGA} & \multicolumn{1}{c}{F1} & \multicolumn{1}{c}{JGA} & \multicolumn{1}{c}{F1} \\

\midrule[0.8pt]
\multicolumn{5}{c}{LLaMA-30B~\citep{llama}}                                                          \\
\midrule[0.8pt]
IC-DST (SBERT)            & $25.41$\tiny$\pm1.82$  & $77.82$\tiny$\pm2.16$ & $26.01$\tiny$\pm2.17$ & $79.01$\tiny$\pm2.52$        \\
SM2 &  $22.86$\tiny$\pm1.35$ & $74.73$\tiny$\pm1.95$ &$23.46$\tiny$\pm1.80$ & $75.78$\tiny$\pm2.41$ \\
GTR-T5 &   $25.10$\tiny$\pm0.33$ & $68.42$\tiny$\pm1.93$    &$19.94$\tiny$\pm2.40$   &$68.90$\tiny$\pm2.22$     \\
Jina &   $22.51$\tiny$\pm0.92$ & $72.31$\tiny$\pm1.01$  & $22.42$\tiny$\pm1.18$& $72.95$\tiny$\pm0.93$               \\
\midrule[0.5pt]
Sum. + GTR-T5 & $26.06$\tiny$\pm0.47$ & $78.55$\tiny$\pm0.35$ & $26.75$\tiny$\pm0.93$ & $78.55$\tiny$\pm0.35$                  \\
Sum. + Jina  & $25.10$\tiny$\pm0.33$ &  $78.07$\tiny$\pm0.54$ &$25.81$\tiny$\pm1.02$ & $78.98$\tiny$\pm0.66$   \\
\midrule[0.5pt]
\textbf{CONVERSE}  & $\textbf{27.35}$\tiny$\pm0.77$  & $\textbf{79.75}$\tiny$\pm0.95$   & $\textbf{28.23}$\tiny$\pm1.58$ & $\textbf{80.45}$\tiny$\pm0.55$                                 \\
\bottomrule
\end{tabular}
}
\vspace{-0.1in}
\caption{\small JGA and F1 of LLaMA-30B with 100 labeled conversations.}
\label{tab:main-exp-large}
\vspace{-0.2in}
\end{table}

\paragraph{Main Results} The DST results are shown in Table~\ref{tab:main-exp} and Table~\ref{tab:main-exp-large}. The first set of comparisons is between conversation retrieval with and without explicit query/key generation. We observe that using the summary as search keys/queries significantly improves the end-to-end (E2E) results, when evaluated with the same off-the-shelf retriever (GTR-T5 or Jina). The result is slightly behind IC-DST which fine-tunes the retriever with dialogue state information in the key. However, after introducing the distilled CONVERSE model, we achieve much better E2E results than all baselines. Although our motivation of distilling a conversation encoder is to reduce the inference cost, it turns out that the light-weight model is also helpful in E2E performance. We hypothesize that the improved performance brought by CONVERSE is attributed to two factors. The first and foremost is that we leverage a dual encoder architecture to optimize the matching between conversation-summary pairs. This suggests that the retrieval component is optimized for the task-specific keys and values. A secondary explanation of the performance gain is that the conversation encoder avoids error propagation in explicit summary decoding and re-embedding.
It should be noted that the above findings are consistent across different datasets (MultiWOZ 2.1/2.4) and language models (GPT-Neo and LLaMA-7B/30B).

\begin{table}[t]
\resizebox{0.48\textwidth}{!}{\begin{tabular}{lcc}
\toprule
                     & \multicolumn{2}{c}{\textbf{JGA}}                                   \\
\midrule
Model                & \multicolumn{1}{c}{MWZ-2.1} & \multicolumn{1}{c}{MWZ-2.4} \\
\midrule
DS2 + BART-Large           & $7.60$\tiny$\pm2.17$                     & $5.86$\tiny$\pm4.52$                     \\
DS2 + T5-Large           & $17.71$\tiny$\pm1.84$                     & $19.08$\tiny$\pm1.23$                     \\
\midrule
\textbf{CONVERSE} + LLaMA-7B  & ${19.33}$\tiny$\pm0.91$  & ${20.35}$\tiny$\pm1.03$                     \\
\textbf{CONVERSE} + LLaMA-30B & $\textbf{27.35}$\tiny$\pm0.77$                     & $\textbf{28.23}$\tiny$\pm1.58$                     \\
\bottomrule
\end{tabular}
}
\vspace{-0.1in}
\caption{\small Comparison against few-shot finetuning methods.}
\label{tab:few-shot}
\end{table}
\paragraph{Comparison Against Few-Shot Finetuning}
Recently, \citet{few-shot-vs-icl} have shown that few-shot fine-tuning outperforms in-context learning in some settings, which makes people wonder how few-shot fine-tuning behaves with 100 labeled dialogues for dialogue state tracking tasks. To answer this question, we compare our method, CONVERSE, against one of the strongest few-shot fine-tuning methods, DS2 \citep{few-shot-dst-summarization}, using BART \citep{bart} and T5 \citep{T5} language models. As shown in Table \ref{tab:few-shot}, our method, CONVERSE, outperforms the few-shot fine-tuning method, DST. Note that the BART-based DS2 severely overfits to the small labeled dataset and the T5-based model performs worse than the in-context learning method, even though T5 model is pretrained on an additional large-scale labeled dialogue summarization dataset, SAMSum \citep{samsum}.

\begin{table}[t]
\centering
\resizebox{0.48\textwidth}{!}{\begin{tabular}{lllll}
\toprule
                       & \multicolumn{2}{c}{\textbf{MultiWOZ 2.1}}                 & \multicolumn{2}{c}{\textbf{MultiWOZ 2.4}}                 \\
\cmidrule(lr){2-3} \cmidrule(lr){4-5}
Model                  & \multicolumn{1}{c}{JGA} & \multicolumn{1}{c}{F1} & \multicolumn{1}{c}{JGA} & \multicolumn{1}{c}{F1} \\
\midrule[0.8pt]

\multicolumn{5}{c}{LLaMA-7B~\citep{llama}}                                                                                            \\
\midrule[0.8pt]
IC-DST (SBERT)                          &   $12.52$\tiny$\pm0.68$            &   $62.11$\tiny$\pm0.38$       & $12.43$\tiny$\pm0.09$                 &   $62.45$\tiny$\pm0.87$                     \\

\midrule
\textbf{CONVERSE}           &    $\textbf{14.05}$\tiny$\pm0.58$  &   $\textbf{63.37}$\tiny$\pm1.53$  & $\textbf{14.23}$\tiny$\pm0.48$    &   $\textbf{64.18}$\tiny$\pm1.47$                     \\
\bottomrule
\end{tabular}
}
\vspace{-0.1in}
\caption{\small Out-of domain generalization using 100 labeled conversations with LLaMA-7B.}
\label{tab:ood}
\vspace{-0.2in}
\end{table}
\paragraph{Out-of Domain Generalization}
To verify our hypothesis that our unsupervised retriever CONVERSE generalizes better to unseen domain than supervised methods, we hold out the hotel domain from the MultiWOZ dataset and train the retrievers, IC-DST and CONVERSE on the remaining four domains: train, restaurant, taxi, and attraction. Then we evaluate the performance of the few-shot in-context learning with the retrievers on test examples from the unseen domain, hotel. As shown in Table~\ref{tab:ood}, our model CONVERSE outperforms IC-DST by a large margin, which empirically validates that our unsupervised retriever generalizes better to unseen domain than the supervised one.

\subsection{Ablation Study \label{ablation}}
\begin{figure}[t]
    \centering
\includegraphics[width=0.9\linewidth]{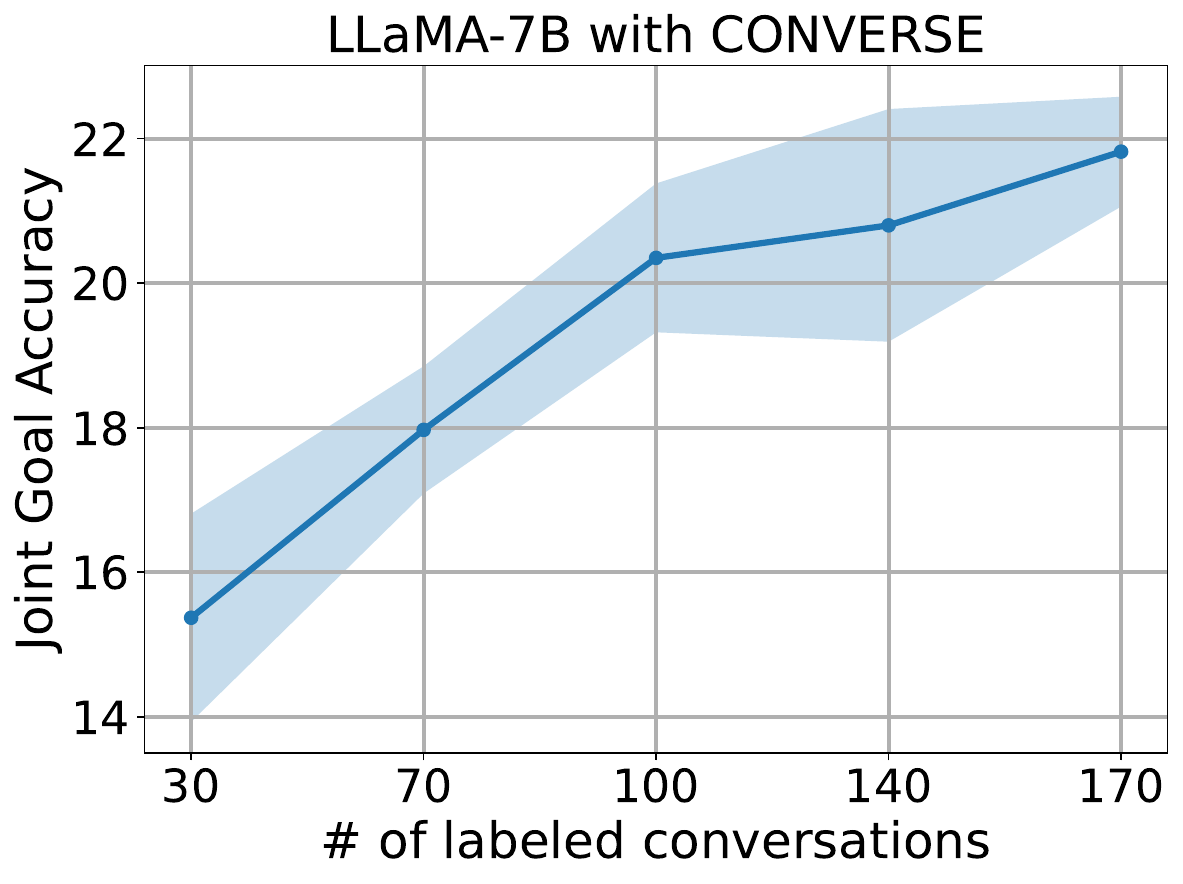}
    \vspace{-0.1in}
    \caption{\small JGA of LLaMA-7B with CONVERSE as a function of the number of labeled data.}
    \vspace{-0.1in}
    \label{fig:num-labels}
\end{figure}

\paragraph{Size of Support Set} 
We empirically study the size of the support set (labeled dialogues) in the conversation retrieval task. Notably, a smaller support set requires less annotation effort from the domain owner, placing more emphasis on generalization to unseen dialogue structures. In contrast, a larger support set contradicts the fundamental motivation behind few-shot learning, but it is likely to improve the E2E accuracy, as more test dialogue structures are observable from the exemplars.  
In Figure~\ref{fig:num-labels}, we plot the JGA of LLaMA-7B  with CONVERSE on varying sizes of the support set constructed from  MultiWOZ. As we expect, the JGA increases as the number of labeled conversation increases, even though we do not fine-tune the retriever with any labeled conversations.

\begin{figure*}
\vspace{-0.1in}
	\begin{subfigure}{.9\textwidth}
		\centering
		\includegraphics[width=\linewidth]{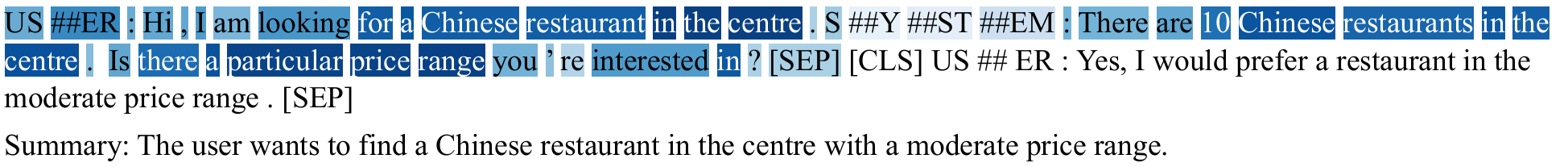}
  \captionsetup{justification=centering,margin=0.5cm}
    \vspace{-0.1in}		
  \caption{}
		\label{fig:vis-1}
\vspace{-0.1in}
 \end{subfigure}

  \rule{\textwidth}{0.5pt}		   
  
    \begin{subfigure}{.9\textwidth}
		\centering
		\includegraphics[width=\linewidth]{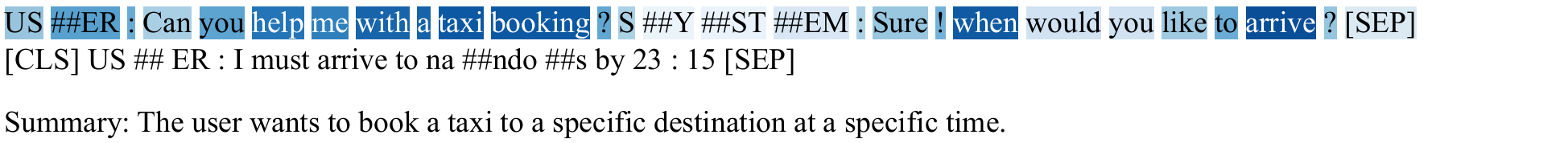}
		\captionsetup{justification=centering,margin=0.5cm}
        \vspace{-0.1in}				
        \caption{}
		\label{fig:vis-2}
\vspace{-0.1in}	
 \end{subfigure}

\hrulefill

        \begin{subfigure}{.9\textwidth}
		\centering
		\includegraphics[width=\linewidth]{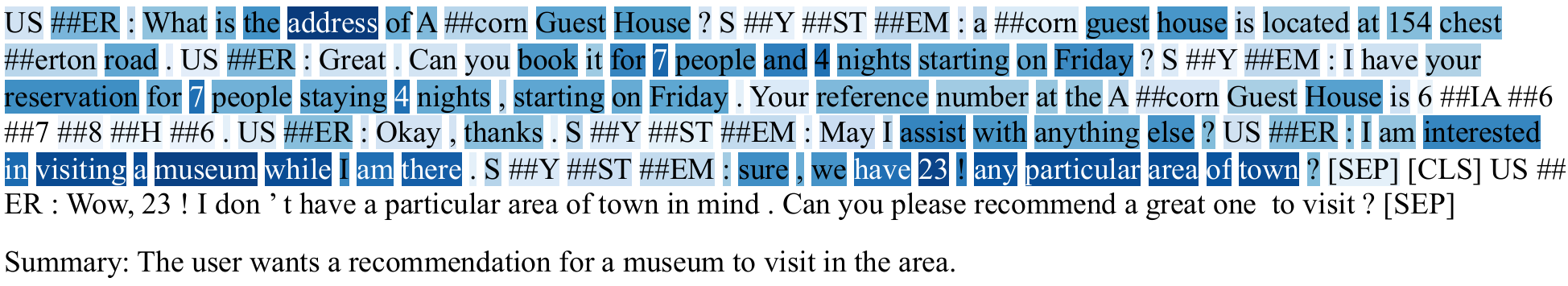}
		\captionsetup{justification=centering,margin=0.5cm}
    \vspace{-0.1in}		
  \caption{}
		\label{fig:vis-3}
	\end{subfigure}
 \vspace{-0.1in}
 \caption{\small\textbf{Visualization of importance scores.} Tokens with darker blue gets larger weights based on the latest user utterance.}
\label{fig:visualize}
\vspace{-0.1in}
\end{figure*}

\paragraph{Summary vs. state delta}
The conversation summary we adopt in this work concludes the user's current intent when the dialogue takes place. A limitation is that the summary does not directly highlight the state delta carried by the latest user input. As a remedy, we consider a multi-key and query retrieval setup, where we use both the summary and the latest user input as search keys and queries. More specifically, we first retrieve 20 dialogues with CONVERSE and re-rank the 20 dialogues based on the similarity of the latest utterance between the test sample and the support examples, using the pre-trained GTR-T5-Large.
As shown in Table~\ref{tab:retriever}, re-ranking with the latest user utterance yields marginal performance gains. In future work, we aim to explore a better way of summarizing the conversation structure that reflects both the joint intent and the latest user input.

\begin{table}[t]
    \centering
\resizebox{0.45\textwidth}{!}{\begin{tabular}{lcc}
    \toprule
    \multicolumn{3}{c}{\textbf{JGA}} \\
    \midrule[0.5pt]
    Model     &  {\textbf{MultiWOZ 2.1}} & \textbf{MultiWOZ 2.4}\\
    \midrule[0.5pt]
    \textbf{CONVERSE + Rerank} &  ${19.86}\pm1.22$ & ${20.65}\pm1.28$ \\
    \textbf{CONVERSE} &  ${19.33}\pm0.91$ & ${20.35}\pm1.03$ \\
    \bottomrule
    \end{tabular}
}
\vspace{-0.1in}
\caption{\small Ablations of different ranking with LLaMA-7B.}
\label{tab:retriever}
\vspace{-0.2in}
\end{table}

\subsection{Qualitative Results \label{Qualitative}}

\paragraph{Visualization of history grounding} 
As described in~\eqref{eq:weight}, CONVERSE softly retrieves conversation history based on the latest user utterance. Specifically, the network $g_\phi$ outputs a relevance score between 0 and 1 for each token of the conversation history. 
In Figure~\ref{fig:visualize}, we visualize this relevance score of each token in the history. The tokens with darker blue color indicates a higher weight, which are considered to be more relevant to the latest input. 

The examples in Figure~\ref{fig:visualize} shows that the model successfully focuses on relevant part of history. For the first example in Figure~\ref{fig:vis-1}, the user wants to search for a Chinese restaurant in the center with moderate price range. The model assigns large weights to the tokens related to ``Chinese'', ``center'', and ``price''. Similarly, the tokens relevant to booking a taxi gets larger weights in Figure~\ref{fig:vis-2}. For the last example in Figure~\ref{fig:vis-3}, the model pays attention to the tokens related to a museum and ignores many  irrelevant ones. 

\begin{table}[t]
    \centering
    \resizebox{0.482\textwidth}{!}{\begin{tabular}{l}
    \toprule
    \multicolumn{1}{c}{\textbf{Conversation}} \\
    \midrule[0.5pt]
    \textbf{USER}: I need some tourist information please. I need to  \\
    know about a hotel called the Arbury lodge guest house.\\ 
    \textbf{SYSTEM}: The Arbury lodge guest house is in the north area \\
    and has a moderate price range. $\cdots$ \\
    \textbf{USER}: I would like to book a stay for 3 people for\\
    2 nights starting from Tuesday.\\
    \textbf{USER}: I am also looking to eat somewhere expensive, \\
    in the south area of town. \\
    \multicolumn{1}{c}{\small$\vdots$} \\
    \textbf{USER}: I will also need a taxi , please. \\
    \textbf{SYSTEM}: Where would you like your taxi to pick you up \\
    and drop you off? \\
    \textbf{USER}: I want to be picked up at the hotel and dropped off \\
    at the restaurant. \\
    \midrule[0.5pt]   
    \color{ourdarkblue}{\textbf{Summary}: The user wants to book a taxi to be}  \\
    \color{ourdarkblue}{picked up at a specific location and dropped off at another.} \\
    \bottomrule
    \end{tabular}}
    \vspace{-0.1in}
    \caption{\small The LLM successfully summarize the conversation based on the latest user utterance.}
    \label{tab:summary-1}
\vspace{-0.2in}
\end{table}

\paragraph{Human Evaluation on Conversation Summarizer}
The success of CONVERSE is highly dependent on the output quality of the conversation summarizer, which are used as labels for encoder distillation. 
We conduct human evaluation of 135 summaries generated by the conversation summarizer, namely \texttt{gpt-3.5-turbo}. Specifically, three human judges are asked to assess whether the generated summaries are consistent with the instructions in the prompt in Table~\ref{tab:summary-prompt}. The results indicate that 90.3\% of the 135 summaries are deemed consistent with the given prompt.

\begin{table}[t]
    \centering
    \resizebox{0.4\textwidth}{!}{\begin{tabular}{l}
     \toprule
     \multicolumn{1}{c}{\textbf{Conversation}} \\
     \midrule[0.5pt]
     \multicolumn{1}{c}{\small$\vdots$} \\
    \textbf{SYSTEM}: Booking was successful.  \\
    The table will be reserved for 15 minutes.  $\cdots$ \\
    \textbf{USER}: Great. 1 more thing. Can you book a taxi \\
    between the 2 places? I would like to arrive  at the \\ 
    restaurant  in time for my reservation \\
    \midrule[0.5pt]
    \color{ourdarkblue}{\textbf{Summary}: The user wants to book a taxi to travel} \\
    \color{ourdarkblue}{between two specific locations.} \\
    \bottomrule
    \end{tabular}}
    \vspace{-0.1in}
    \caption{\small A failure case of summarization with the LLM.}
    \label{tab:summary-2}
\vspace{-0.2in}
\end{table}

Examples of the generated summaries are shown in Table~\ref{tab:summary-1} and ~\ref{tab:summary-2}. For the first example, the model generates the summary about booking a taxi. It is noteworthy that the model focuses on the latest user utterance while disregarding previous user requests for hotel and restaurant reservation. For the second example, the model misses out on the arrival time for generating the summary. 
Identification and correction of such errors are topics we will explore in future work. We include more examples in Appendix~\ref{app:summ}.
\begin{table*}[t]
    \centering
    \resizebox{0.99\textwidth}{!}{\begin{tabular}{l|l}
    \toprule
          \multirow{3}{*}{\begin{tabular}[c]{@{}c@{}}Target\\ Conversation\end{tabular}} & \multicolumn{1}{c}{\small$\vdots$} \\
          &\textbf{SYSTEM}: There are 9 Indian restaurants in centre what price range do you want? \\
          & \textbf{USER}: I am looking for expensive Indian food. \\
         \midrule[0.5pt]
         {Gold Label} &  restaurant-food: indian, restaurant-pricerange: expensive \\
         \midrule[0.5pt]
         {Prediction} & restaurant-food: indian, restaurant-pricerange: expensive \\
    \midrule[0.5pt]
    \multirow{6}{*}{Exemplar \#1} &  \multicolumn{1}{c}{\small$\vdots$}\\
    & \textbf{USER}: I am also looking to eat somewhere expensive, in the south area of town.\\
    & \textbf{SYSTEM}: There are 2 Chinese, 1 Indian , 1 Italian, and 1 Mexican restaurants. Which of those would you like?\\ 
    & \textbf{USER}: I would like the Italian place please.  \\
    & \color{ourdarkblue}{\textbf{Summary}: The user wants to find a restaurant in a specific area with a certain price range and cuisine.} \\
    & \textbf{Label}: restaurant-food: italian. \\
    \midrule[0.5pt]
    \multirow{6}{*}{Exemplar \#2} &  \multicolumn{1}{c}{\small$\vdots$}\\
    & \textbf{USER}: Actually, I also need a moderate priced restaurant in the same area.\\
    & \textbf{SYSTEM}: I can find a few that meet that criteria, would you like Indian, or Italian food?\\ 
    & \textbf{USER}: Well, everyone said it's my choice, so I think I would like Italian.  \\
    & \color{ourdarkblue}{\textbf{Summary}: The user wants to find a moderate priced restaurant in a specific area with a specific cuisine.} \\
    & \textbf{Label}: restaurant-food: italian. \\
    \midrule[0.5pt]
    \multirow{4}{*}{Exemplar \#3} &  \multicolumn{1}{c}{\small$\vdots$}\\
    & \textbf{SYSTEM}: Good news I was able to get this for you. Reference i4dxhdjl. Can I help you find other things to do in the area as well ?\\ 
    & \textbf{USER}: I am also looking to eat somewhere expensive, in the south area of town.  \\
    & \color{ourdarkblue}{\textbf{Summary}: The user wants to find a restaurant with an expensive price range in a specific area of town.} \\
    & \textbf{Label}: restaurant-pricerange: expensive, restaurant-area: south \\
    \bottomrule
    \end{tabular}
    }
    \vspace{-0.1in}
    \caption{\small Given the target conversation, we show the top 3 most similar examples retrieved by our model CONVERSE.}
    \label{tab:retrieval-example}
    \vspace{-0.1in}
\end{table*}

\paragraph{Retrieved Exemplars} In Table~\ref{tab:retrieval-example}, we show the top three most similar examples retrieved by CONVERSE. In this example, the user asks to find an expensive Indian restaurant and a retriever needs to retrieve conversations about a restaurant. Indeed, our CONVERSE retriever assigns high similarity scores to pairs of the target conversation and summaries about finding a restaurant.  Note that the language model (LLaMA-7B) with in-context learning successfully generalizes to decode  test slot values from the exemplars, though  the retrieved exemplars consist of values for food or price range, which are  different from the target conversation.
\section{Related Work}

\paragraph{Dialog State Tracking}
Most of existing works on DST train a supervised model with large-scale labeled datasets~\citep{dst-1, dst-2, dst-3, dst-4, dst-5, dst-6, dst-7, dst-8, dst-9, dst-10, dst-11, dst-12}. However, a supervised model does not scale well to new domains or annotation schemas. To address the problem, several recent works explore few-shot DST~\citep{few-shot-dst-1, few-shot-dst-2, few-shot-dst-3, few-shot-dst-4, few-shot-dst-5, few-shot-dst-6}. Most related are the works of~\citet{ic-dst, smc-2}, who adopt in-context learning with LLM for dialog state generation. The work demonstrated the few-shot generalization ability of LLM applied to DST without parameter updates, but the dialog retriever is still fine-tuned with in-domain data.

Another work related to ours is  \citet{few-shot-dst-summarization}, which formulates DST as a summarization task. 
The authors train a T5 language model to decode text summaries, which are then transformed into dialog states with heuristic rules. Different from their work, we do not aim to alter the target of DST as summaries but rather our goal is to enable effective conversation retrieval.

\paragraph{Retrieval}
Our work mainly focuses on retrieving relevant conversations for in-context learning~\citep{in-context-retrieval}. There is a vast number of papers~\citep{dpr, colbert, contriever, colbert-2} proposing neural network based retrievers which encode queries and keys into low dimensional vectors and compute similarities between them. \citet{ic-dst, smc-2} propose to utilize slots and values to represent a long history of conversation for retrieval. However, in order to train the retriever, their approaches require labeled dialogue data to construct positive and negative conversations for each query conversation. Recently, \citet{abs-retrieval} retrieve texts based on abstract descriptions generated by a LLM.

\section{Conclusion}
The contribution of this work is twofold. First, we proposed an effective way of retrieving conversations in LLM-based DST with conversation summaries as search keys and queries. We then improved the efficiency of the retrieval system by distilling a conversation encoder capable of embedding a conversation into a vector space similar to its summary.
This eliminates the cost of decoding an actual summary for each test sample during inference.
We validated  our CONVERSE encoder for LLM-based DST in a real few-shot setting with 100 conversations in the support set. Results showed that CONVERSE consistently improved both the efficiency and the performance of few-shot DST when using different LLMs, outperforming previous LLM-based DST baselines that rely on annotated dialogues for retriever fine-tuning.

\paragraph{Acknowledgement}
We would like to express our sincere gratitude to Dominik Wagner and Minki Kang for the numerous helpful pieces of feedbacks.

\bibliography{reference}
\bibliographystyle{acl_natbib}
\clearpage

\appendix

\section*{Appendix}

\begin{table*}[ht!]
    \centering
    \resizebox{0.95\textwidth}{!}{\begin{tabular}{l}
    \toprule
You are shown a conversation between a virtual assistant on a phone and a user.  You have to summarise \\
what the user wants at this particular point of the conversation. 
You summary should contain the user  \\
intent  and the slots he mentioned. 
However, the summary should be a delexicalized abstrast sentence, \\
which means   it should not contain actual slot values. 
Note that it is possible that not all conversation \\
history is relevant and you need to summarise based on what is relevant to the most recent user turn.  \\
If the user does not 
have a goal   at this point or his goal gets completed by the system,  \\
just summarize that  ``The user wants nothing more''. \\
\\
  \color{ourdarkblue}{<fictional$\_$example>} \\

  USER: make an alarm for 6 \\
  SYSTEM: I have created an alarm at 6 \\
  USER: Also, send a message to my wife \\
  SYSTEM: What would you like the message to say? \\
  USER: ehm... happy birthday \\
  SYSTEM: I can do that. What message service do you want to use \\
  USER: whatsapp \\
\\
  What does the user want at this point in the conversation? \\
  The user wants to send a message to a recipient with a given text using a specified app \\
  \color{ourdarkblue}{</fictional$\_$example>} \\
\\
  \color{ourdarkblue}{<fictional$\_$example>} \\

  USER: make an alarm for 6 \\ 
  SYSTEM: I have created an alarm at 6 \\ 
  USER: thanks you and goodbye \\ 
\\
  What does the user want at this point in the conversation? \\
  The user wants nothing more \\ 

  \color{ourdarkblue}{</fictional$\_$example>} \\ \\

  Now it's your turn. \\

  \color{ourdarkblue}{$\{$test$\_$example$\}$} \\
  What does the user want at this point in the conversation? \\
  \bottomrule
    \end{tabular}
    }
    \vspace{-0.1in}
    \caption{\small Prompt with two exemplars for summarizing a conversation.}
    \label{tab:summary-prompt}
\end{table*}
\section{Prompt for Summarization}\label{app:prompt}
In Table~\ref{tab:summary-prompt}, we provide an instruction and two examples that specify how to summarize a conversation based on the latest user utterance. The first example demonstrate how to ignore irrelevant history --- request for setting an alarm and focus on sending a message. The second example is to generate ``The user wants nothing more'' instead of summarizing the conversation.

\section{Instruction for Human Evaluation}
\begin{figure*}
    \centering
    \includegraphics[width=0.7\linewidth]{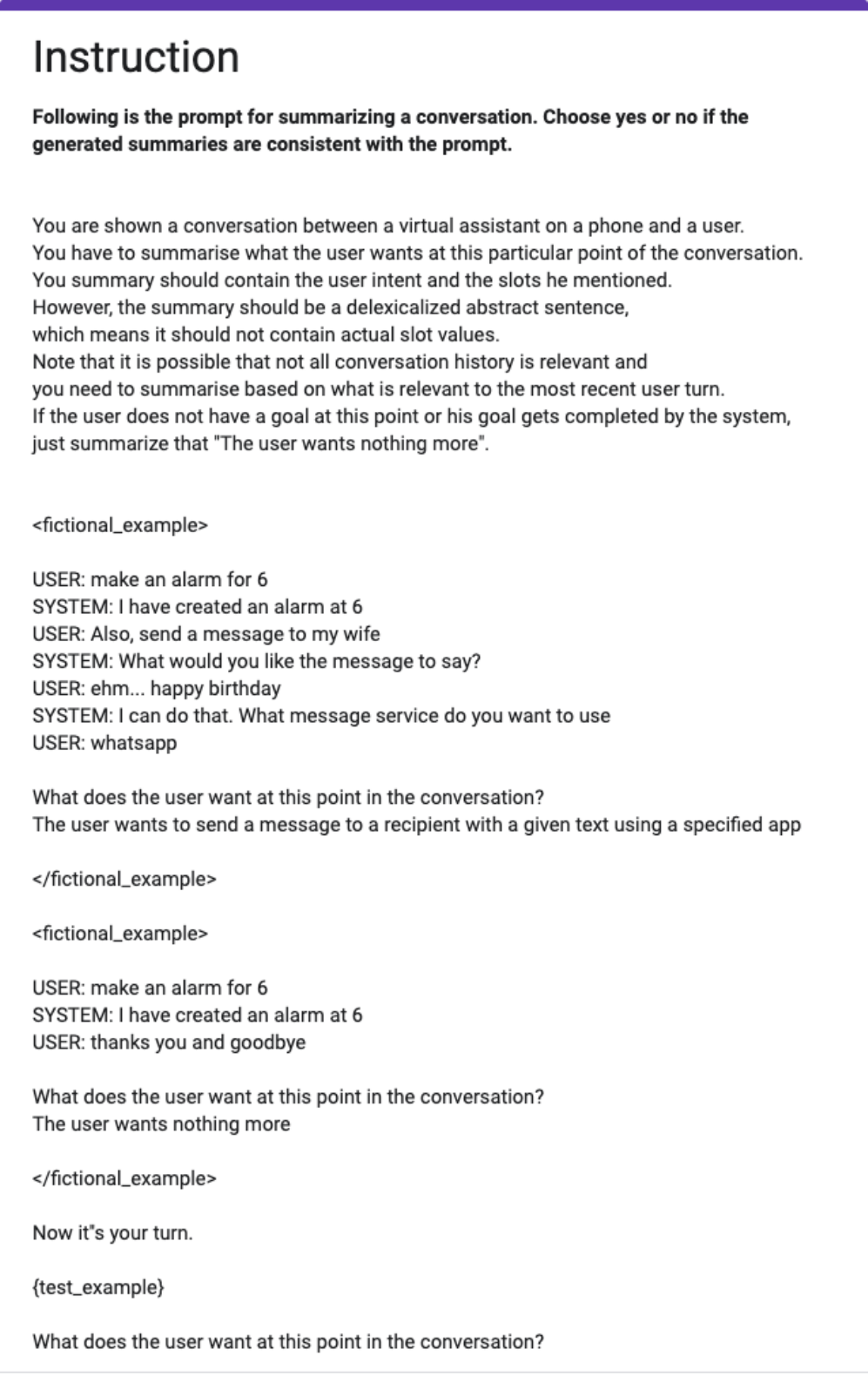}
    \caption{\small A screenshot of the instruction for human evaluation on summaries generated by gpt-3.5-turbo.}
    \label{fig:instruction}
\end{figure*}
We conduct human evaluation of 135 summaries generated by the conversation summarizer, \texttt{gpt-3.5-turbo}. As shown in Figure~\ref{fig:instruction}, three human judges are given an instruction for the evaluation and they are asked to assess whether the generated summaries are consistent with the prompt specified in Table~\ref{tab:summary-prompt}.

\begin{table*}[t]
    \centering
    \resizebox{0.95\textwidth}{!}{\begin{tabular}{l}
    \toprule
    \multicolumn{1}{c}{\textbf{Conversation}} \\
    \midrule[0.5pt]

    \textbf{USER}: I am looking for a hotel in the east , and i would like free wifi please. \\ 
    \textbf{SYSTEM}: I have found express by holiday inn Cambridge. 1517 Norman way, Coldhams business park\\
    \textbf{USER}: Does that hotel include free parking and have 2 stars? \\
    \textbf{SYSTEM}: No, thank you , that s all I need. \\
    \textbf{USER}: Sure thing what's the area and or name? \\
    \textbf{SYSTEM}: Thank you for contacting Cambridge Towninfo centre. Have a nice trip! Good Bye! \\
    \textbf{USER}: Actually, I also need a moderate priced restaurant in the same area  \\
    \midrule[0.5pt]   
    \color{ourdarkblue}{\textbf{Summary}: The user wants to find a moderate priced restaurant in a specific area.}  \\
    \bottomrule
    \end{tabular}}
    \vspace{-0.1in}
    \caption{\small Example of conversation summary.}
    \label{tab:summary-3}
\end{table*}

\begin{table*}[ht]
    \centering
    \resizebox{0.95\textwidth}{!}{\begin{tabular}{l}
    \toprule
    \multicolumn{1}{c}{\textbf{Conversation}} \\
    \midrule[0.5pt]
    \textbf{USER}: I need some tourist information please. I need to know about a hotel called the Arbury lodge guest house. \\ 
    \textbf{SYSTEM}: The arbury lodge guest house is in the north area and has a moderate price range. It is 4 stars and has \\ free parking and internet.\\
    \textbf{USER}: I would like to book a stay for 3 people for 2 nights starting from Tuesday. \\
    \textbf{SYSTEM}: Good news I was able to get this for you. Reference i4dxhdjl. \\ Can I help you find other things to do in the area as well?\\
    \textbf{USER}: I am also looking to eat somewhere expensive, in the south area of town. \\
    \textbf{SYSTEM}: there are 2 Chinese, 1 Indian, 1 Italian, and 1 Mexican restaurants. Which of those would you like? \\
    \textbf{USER}: I would like the Italian place please.\\
    \textbf{SYSTEM}: Frankie and Benny s is an expensive restaurant which is expensive would you like for me to book it? \\
    \textbf{USER}: yeah , book for 3 people on tuesday at 18:30 and give me the reference number \\
    \midrule[0.5pt]   
    \color{ourdarkblue}{\textbf{Summary}: The user wants to book a table at an expensive Italian restaurant for 3 people at a specific date }\\
    \color{ourdarkblue}{and time and receive a reference number.}  \\
    \bottomrule
    \end{tabular}}
    \vspace{-0.1in}
    \caption{\small Example of conversation summary.}
    \label{tab:summary-4}
\vspace{-0.1in}
\end{table*}

\begin{table*}[ht]
    \centering
    \resizebox{0.95\textwidth}{!}{\begin{tabular}{l}
    \toprule
    \multicolumn{1}{c}{\textbf{Conversation}} \\
    \midrule[0.5pt]
    \textbf{USER}: I would like to get to Cambridge by train, but I am afraid I need to travel on Sunday. Will that be possible? \\ 
    \textbf{SYSTEM}: From where are you departing?\\
    \textbf{USER}: I am leaving from Broxbourne. \\
    \textbf{SYSTEM}: There are 19 trains running from Broxbourne to Cambridge on Sunday.  \\
    \midrule[0.5pt]   
    \color{ourdarkblue}{\textbf{Summary}: The user wants to find a train from a specific location to a specific destination}\\ 
    \textcolor{ourdarkblue}{arriving before a certain time.}\\
    \bottomrule
    \end{tabular}}
    \vspace{-0.1in}
    \caption{\small Example of conversation summary.}
    \label{tab:summary-5}
\vspace{-0.1in}
\end{table*}

\begin{table*}[ht]
    \centering
    \resizebox{0.95\textwidth}{!}{\begin{tabular}{l}
    \toprule
    \multicolumn{1}{c}{\textbf{Conversation}} \\
    \midrule[0.5pt]
    \textbf{USER}: Please help me reserve a taxi from saint catharine s college\\ 
    \textbf{SYSTEM}: Sure! When would you like to arrive?\\
    \textbf{USER}: I would like to arrive by 01:15. \\
    \textbf{SYSTEM}: Great! And what is your destination?  \\
    \textbf{USER}: I am going to Loch Fyne. \\
    \midrule[0.5pt]   
    \color{ourdarkblue}{\textbf{Summary}: The user wants to reserve a taxi from a specific location to a specific destination at a specific time.}\\ 
    \bottomrule
    \end{tabular}}
    \vspace{-0.1in}
    \caption{\small Example of conversation summary.}
    \label{tab:summary-6}
\vspace{-0.1in}
\end{table*}

\begin{table*}[ht]
    \centering
    \resizebox{0.95\textwidth}{!}{\begin{tabular}{l}
    \toprule
    \multicolumn{1}{c}{\textbf{Conversation}} \\
    \midrule[0.5pt]
    \textbf{USER}: Are there any concert halls in the west?\\ 
    \textbf{SYSTEM}: I am sorry there are no listings for concert halls in the west. Is there something else \\
    I can assist you with today?\\
    \textbf{USER}: What about colleges in the west  \\
    \textbf{SYSTEM}: There are 5 colleges in that area. 3 have no entrance fees. \\
    Would you like additional information on any of these?  \\
    \textbf{USER}: No thank you. I am looking for information El Shaddai hotel. \\
    \textbf{SYSTEM}: The El Shaddai is a guest house in the centre part of town. It has free internet , parking and has 0 stars. \\ Would you like me to book it for you? \\
    \textbf{USER}: Could you? That would be great. There are 5 of us and we plan to arrive on Thursday. We'd like to stay for 5 nights. \\
    \midrule[0.5pt]   
    \color{ourdarkblue}{\textbf{Summary}: The user wants to book a hotel for a group of 5 people for a specific duration of time.}\\ 
    \bottomrule
    \end{tabular}}
    \vspace{-0.1in}
    \caption{\small Example of conversation summary.}
    \label{tab:summary-7}
\vspace{-0.1in}
\end{table*}

\label{app:summ}
\section{Example of Summaries}
In~\Crefrange{tab:summary-3}{tab:summary-7}, we provide more summaries generated by the LLM, \texttt{gpt-3.5-turbo}.

\end{document}